# One-Dimensional Vector based Pattern Matching

Y. M. Fouda[a,b]

[a] College of Computer Science and Information Technology, King Faisal University, P.O. Box 400, Al-Ahsa 31982, Kingdom of Saudi Arabia.

[b] Mathematics Department, Faculty of Science, Mansoura University, Mansoura 35516, Egypt.

E-mail: yfoudah@kfu.edu.sa

**Abstract:**

Template matching is a basic method in image analysis to extract useful information from images. In this paper, we suggest a new method for pattern matching. Our method transform the template image from two dimensional image into one dimensional vector. Also all sub-windows (same size of template) in the reference image will transform into one dimensional vectors. The three similarity measures SAD, SSD, and Euclidean are used to compute the likeness between template and all sub-windows in the reference image to find the best match. The experimental results show the superior performance of the proposed method over the conventional methods on various template of different sizes.

**1 Introduction**

Pattern matching is an important technique in pattern recognition and image processing. It is used in many applications related to signal processing and machine vision such as object tracking, stereo matching, video compression, image retrieval and image registration. Template matching tries to answer one of the most basic questions about image: Is there a certain object in that image? If so, where? The template is a description of that object, and is used to search the image by computing a difference measure between the template and all possible portions of the image that could match the template: if any of these produces a small difference, then it is viewed as possible occurrence of the object[1].

Various difference measures have different mathematical properties, and different computational properties had been used to find the best match. The most popular similarity measures are the sum of absolute differences (SAD), the sum of squared difference (SSD), and the normalized cross correlation (NCC). Because of SAD and SSD are computationally fast and algorithms are available which make the template search process even faster, many applications of gray-level image matching use SAD or SSD measures to determine the best match. However, these measure are sensitive to outliers and is not robust to variations in the template, such as those that occur at occluding boundaries in the image. However, the NCC measure is more accurate but it is computationally slow. It is more robust than SAD and SSD under uniform illumination changes, so the NCC measure has been widely used in object recognition and industrial inspection such as in [2] and [3]. An empirical study of five template matching algorithms in the presence of various image distortions has found that NCC provides the best performance in all image categories [4].

Many improvements were occurred on SAD and NCC algorithms to get a better complexity. Mahmood and Khan [5] proposed a partial computation elimination technique in which at a particular search location, computations are prematurely terminated as soon as it is found that this location cannot compete already known best-match location. They showed that partial elimination technique may be applied to correlation coefficient by using a monotonic formulation and proposed a basic mode for small template size and an extended mode for medium and larger template. Chen et al. [6] proposed a fast block matching algorithm based on the winner-update strategy, which can significantly reduce the computation and guarantee to find the globally optimal solution. In their algorithm, only the current winner location with a minimal accumulated distortion is considered for updating the accumulated distortion. This updating process is repeated until the winner has gone through all levels in the pyramids that are constructed from the template and all the candidate windows for the distortion calculation.

Wei and Lai [7] proposed a fast pattern matching algorithm based on NCC criterion by combing adaptive multilevel partition with the winner update scheme to achieve very efficient search. This winner update scheme is applied in conjunction with an upper bound for the cross correlation derived from Cauchy-Schwarz inequality. To apply the winner update scheme in an efficient way, they partition the summation of cross correlation into different levels with the partition order determined by the gradient energies of the partitioned regions in the template. Thus, this winner update scheme in conjunction with the upper bound for NCC can be employed to skip unnecessary calculation. Alsaade and Fouda [8] used Cellular Automata with Rule 170 (CA-R170) on the images after converting it into binary images. Your technique based on eliminating some of the undesirable area in the binary reference images and their corresponding binary template images. Essannouni et. al. [9] proposed a frequency algorithm to speed up the process of SAD matching using Fast Fourier transforms (FFT). They introduced an approach to approximate the SAD metric by cosine series which can be expressed in correlation terms. These latter can be computed using FFT algorithms. Alsaade and Fouda [10] proposed a matching algorithm based on SAD as a measure of similarity and pyramid structure. They applied the pyramid concept to obtain a number of levels of original and template images. Then the SAD measure is applied for each level of image from bottom to up to obtain the correct match in the original image.

In this paper, we propose an efficient pattern search algorithm based on Dimension-Reduction approach for images. Dimension-Reduction technique is applied for the template image and all corresponding sub-windows in the reference image. In our approach Dimensions-Reduction technique based on converting the template and all its corresponding sub-windows in the reference from 2-D into 1-D. The sum of square difference measure was used as a similarity measure to get the template in the reference. The rest of the paper is organized as follows: the proposed algorithm and its complexity analysis is described in section II. Simulation and comparison results for NCC and SAD standards are reported in section III. Then we state conclusions in section IV.

**2 The main contribution**

In this section we introduce description of the proposed method followed by its complexity analysis compared with other four methods.

**2.1 The proposed method**

Our method has been motivated by a need to develop an efficient matching technique so that the detection of objects in a reference image can be effective and fast. The proposed matching algorithm involves three phases. Phase I reduce the amount of data analyzed by transforming 2-D images (template image and sub-windows which has the same size with template in the reference image) into 1-D vector information. This can be done by adding all intensity values for each column in 2-D image see Eq. (1), so we get 1-D vector information. This vector information will be used in the matching process instead of the 2-D image. This transforming reduce the amount of image from $m \times n$ to $n$. Subsequently, this allows the search to be performed with fewer data, while still taking all pixel intensity values into account. Phase II measure the likeness between template image and all possible sub-windows in the reference image. The Euclidean distance or sum of absolute difference or sum of square difference can be used as a similarity measure between 1-D template and all 1-D converted sub-window in the reference. Phase III the decision will be taken based on the similarity values. The sub-window in the reference with minimum similarity value will be the best match for template in the reference.

The basic idea of the proposed template matching depending on converting 2-D template image into 1-D and also the corresponding windows in the reference image over which the template lies. To illustrate the idea suppose that we have a reference image S of size $p \times q$ and template image T of size $m \times n$ where $m < p$ and $n < q$. The problem is to find the best match of template T from the reference image S with minimum distortion.

First the template image $T(i,j)$ converted to 1-D vector NT by the equation

$$NT(i) = \left( \sum_{i=1}^{m} T(i,1), \sum_{i=1}^{m} T(i,2), \sum_{i=1}^{m} T(i,3), \ldots, \sum_{i=1}^{m} T(i,n) \right) \quad (1)$$

where $T(i,j)$ is the pixel value at location $(i,j)$ of the template image.

Secondly, for each pixel $(i,j)$ in the reference image S, $1 \leq i < p-m$ and $1 \leq j < q-n$, we determine a window $W(i,j)$ of size $m \times n$ and all these windows converted into 1-D vector $NW$ by the equation

$$NW(i,j) = \left( \sum_{k=i}^{m+i-1} W(k,j), \sum_{k=i}^{m+i-1} W(k,j+1), \sum_{k=i}^{m+i-1} W(k,j+2), \ldots, \sum_{k=i}^{m+i-1} W(k,n+j-1) \right) \quad (2)$$

where $W(k,j)$ is the pixel value at location $(k,j)$ of the reference image.

Thirdly, the likeness between template image and each corresponding window in the reference are measured by sum of square difference distance between $NT$ and $NW$. All these distances compute and store in new storage $C(i,j)$ where

$$C(i,j) = \left( \sum_{k=i}^{m+i-1} W(k,j) - \sum_{i=1}^{m} T(i,1) \right)^2 + \left( \sum_{k=i}^{m+i-1} W(k,j+1) - \sum_{i=1}^{m} T(i,2) \right)^2 \ldots + \left( \sum_{k=i}^{m+i-1} W(k,n+j-1) - \sum_{i=1}^{m} T(i,n) \right)^2$$
$$\text{where } 1 \leq i < p-m \text{ and } 1 \leq j < q-n. \quad (3)$$

If the positive ordered pair $(\bar{i}, \bar{j})$ be such that $C(\bar{i}, \bar{j})$ is the lowest obtained distance, then return $(\bar{i}, \bar{j})$ as the left upper corner of best template match in our proposed method. Fig. (1) show the concept of the proposed method.

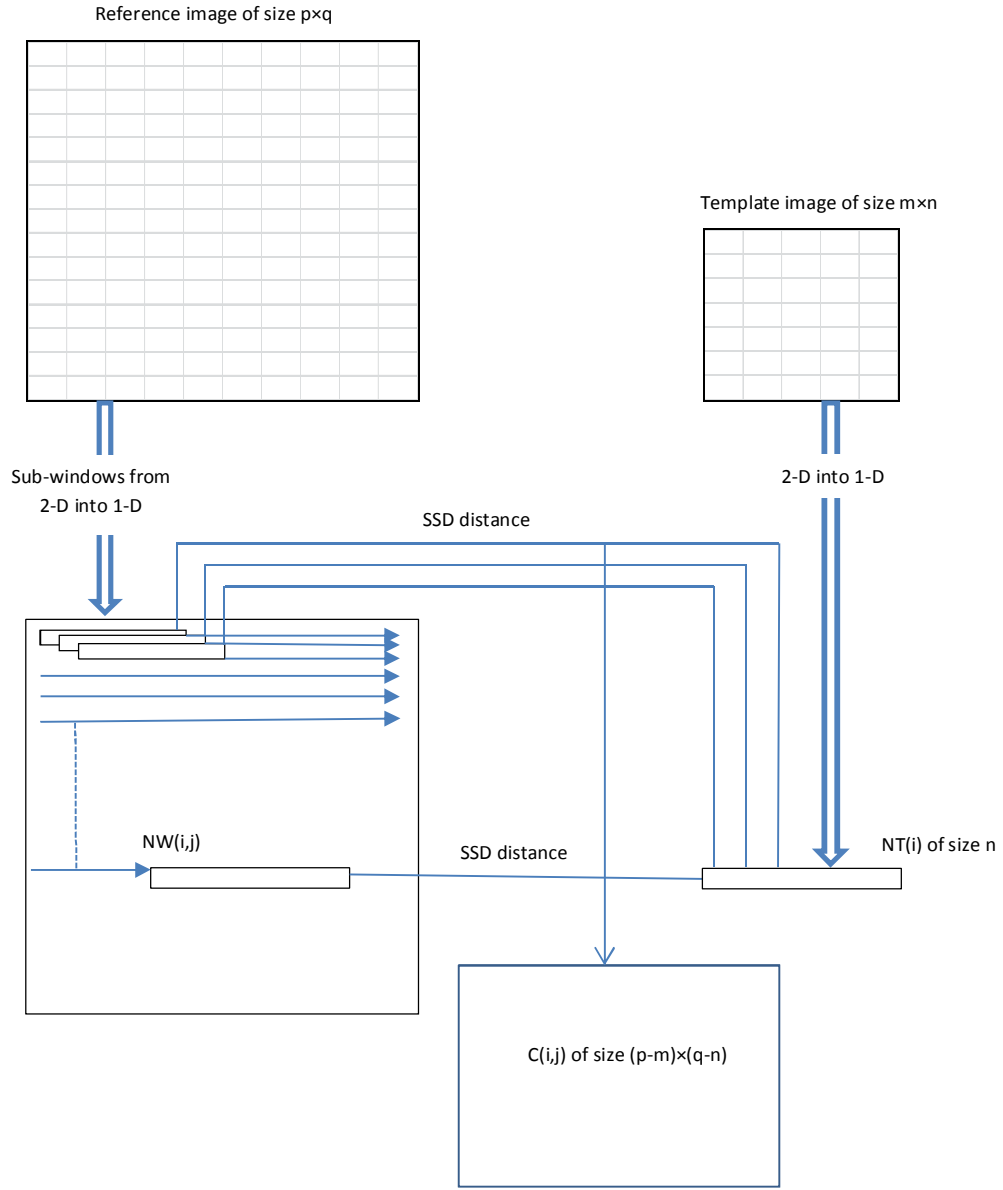

Fig. 1: Illustration of the proposed. The template image is converted into 1-D and also the corresponding windows in the reference, SSD is calculated at every possible location in the reference. The (i,j) location of matrix C with the minimum value is considered as the location of best match.

**2.2 Complexity analysis**

In order to evaluate the efficiency of the proposed algorithm, we discuss the complexity of our algorithm compared with four important methods of template matching NCC, SAD, NCCP, and SADP. For NCC algorithm, the cross correlation coefficient between template image T of size $n \times n$ pixels and an $n \times n$ pixels block in reference image S of size $p \times q$ is given by:

$$\lambda(i,j) = \frac{\sum_{x=1}^{n}\sum_{y=1}^{n}\left(S(i+x, j+y) - \overline{S}(i,j)\right)\left(T(x,y) - \overline{T}\right)}{\sqrt{\sum_{x=1}^{n}\sum_{y=1}^{n}\left(S(i+x, j+y) - \overline{S}(i,j)\right)^2 \sum_{x=1}^{n}\sum_{y=1}^{n}\left(T(x,y) - \overline{T}\right)^2}}, \quad 1 \leq i < (p-n),\ 1 \leq j < (q-n) \tag{4}$$

where

$$\overline{S}(i,j) = \frac{1}{n \times n}\sum_{x=1}^{n}\sum_{y=1}^{n} S(i+x, j+y) \quad \text{and} \quad \overline{T} = \frac{1}{n \times n}\sum_{x=1}^{n}\sum_{y=1}^{n} T(x,y)$$

Direct computation of λ(i,j) require $n \times n = n^2$ (addition/ multiplication) operations at each point *(i,j)* in the reference image where *1≤ i<(p-n), 1≤ j<(q-n)*. Then the operations in Eq. (4) is proportional *to $n^2$(p-n+1)(q-n+1)*. So the computational cost of NCC is O($n^2$(p-n+1)(q-n+1)) which is very time consuming.

For SAD method the sum of absolute difference between the template T of size *n×n* pixels and blocks of sizes *n×n* pixels in the reference image S is given by:

$$SAD(i,j) = \sum_{x=1}^{n}\sum_{y=1}^{n}\left|S(i+x, j+y) - T(x,y)\right| \tag{5}$$

The computation of *SAD(i,j)* requires a number of operations proportional to the template area (*n×n*). These operations are computed for each *(i,j)* in the reference image where *1≤ i<(p-n), 1≤ j<(q-n)*. Then the computational cost for SAD method is O($n^2$(p-n+1)(q-n+1)) the same in NCC algorithm but the SAD method is faster than NCC method because the number of operations in SAD is less than number of operations in NCC for each position *(i,j)* in the reference image about seventy percentage.

The two methods NCCP and SADP can achieve the same estimation accuracy as NCC and SAD while needing much less computation requirement than theses two methods. When the pyramid is applied for NCC and SAD, a sequence of compressed template and reference images are created using:

$$I^k(x,y) = \frac{1}{4}\left(I^{k-1}(2x,2y) + I^{k-1}(2x+1,2y) + I^{k-1}(2x,2y+1) + I^{k-1}(2x+1,2y+1)\right) \tag{6}$$

where $I^k(x,y)$ is the intensity value for the image in the level *k*. The search is conduct using NCC or SAD (Eq. (4) or (5)) with the most compressed template and reference image. The resulting pixel location provides a coarse location of the template pattern in the next lower level of the reference image. Therefore, instead of performing a complete search in the next level, one require to only search a close neighborhood of the area computed from the previous search. This sequence is iterated until the search in the reference image is searched.

In the pyramid concept the complexity of the algorithm depend on the position of template in the reference. If the template coordinates is far from the x-direction and y-direction then applying pyramid outperform the original method. But, if the x-coordinate for template in reference is close to x-axis and/or y-coordinate for template in reference is close y-axis the original method outperform the pyramid concept.

To overcome the problem in pyramid concept and the computational intensive in NCC and SAD we introduced our approach. The computation in Eq.(3) require a number of operations proportional to the length *n* of the converted vector from 2-D into 1-D. These operations are computed for each position *(i,j)* in the reference image where *1≤ i<(p-n), 1≤ j<(q-n)*. Then the computational cost of the proposed method is O(*n(p-n+1)(q-n+1)*) and this justify why the proposed method outperform the others.

## 3 Experimental Results

To measure the efficiency the proposed method was implemented. For the comparing purpose, we also implemented four different algorithms the full-search NCC algorithm, NCC pyramid (NCCP) algorithm, sum of absolute difference (SAD) algorithm, and SAD pyramid (SADP) algorithm. Theses algorithms were implemented in a Matlab 7.0 on a Laptop with an Intel® Core™2 Duo CPU T7500 @ 2.20 GHz and 1.99 GB RAM. Two types of images are used for the testing purpose color images and gray scale images. Greens image of size 300×500 is a representative for color case see fig. 2(a). Lifting-body of size 512×512 is a representative for gray scale case see Fig. 3(a). In our experiments we cropped the templates from the reference (see Fig. 2(b) and Fig. 3(b)) so we know in advance the correct position for template in the reference. The size of these templates varying from 25×25 to 200×200 pixels.

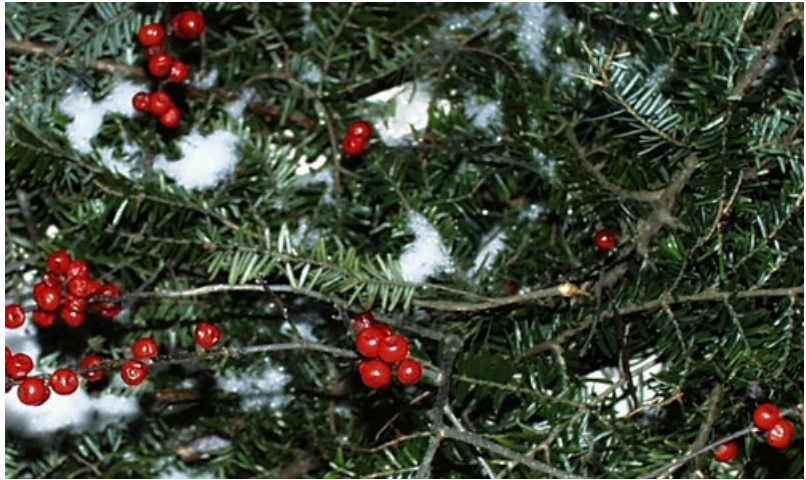

(a)

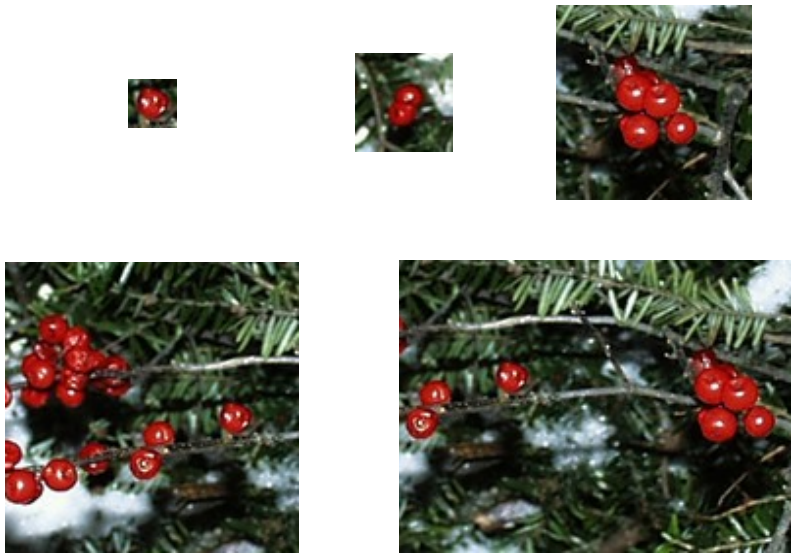

(b)

Fig. 2. (a) Original greens image and (b) The cropped template images form the original with sizes varying from 25x25 to 150x200 pixels.

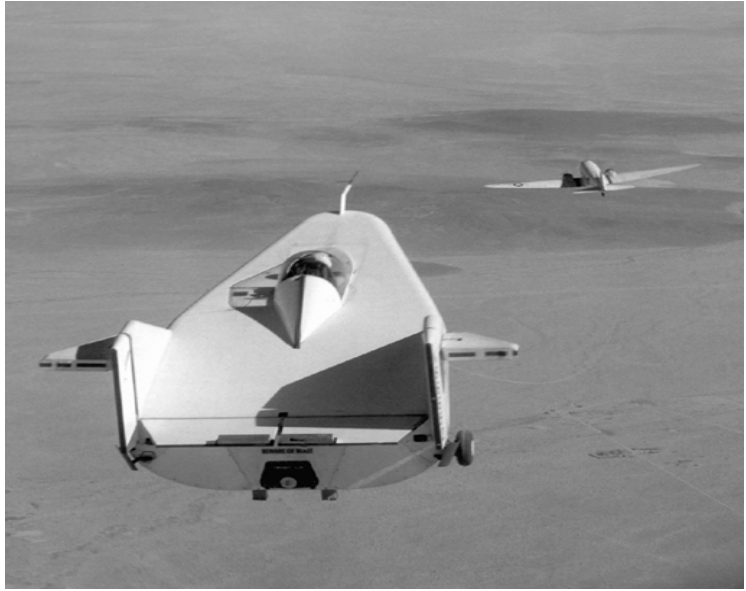

(a)

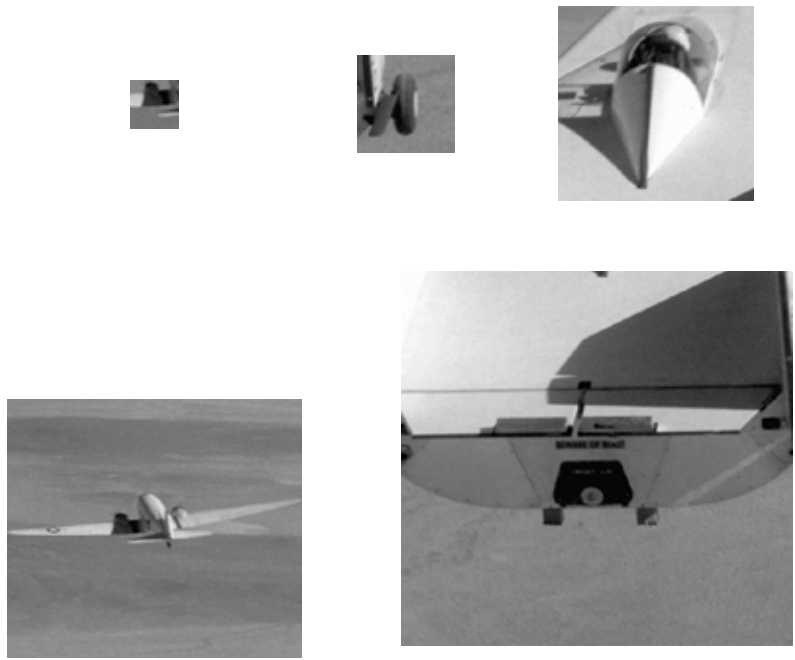

(b)

Fig. 3. (a) Original Lifting-body image and (b) The cropped template images form the original with sizes varying from 25x25 to 200x200 pixels.

In the clean data, the proposed and compared algorithms are guaranteed to find the correct match from the reference image, so we only focus on the comparison of search time required for these algorithms. In the proposed method the similarity between template and sub-windows in the reference image can be measure using more than one similarity function. For example the Euclidean distance or sum of absolute difference or sum of square difference can be used. The mathematical formula for sum of square difference are used in our method (see Eq. (3)). To apply the Euclidean distance and sum of absolute difference functions in our method their mathematical formula is given by the following equations respectively.

$$C(i,j) = \sqrt{\left(\sum_{k=i}^{m+i-1}W(k,j) - \sum_{i=1}^{m}T(i,1)\right)^2 + \left(\sum_{k=i}^{m+i-1}W(k,j+1) - \sum_{i=1}^{m}T(i,2)\right)^2 \ldots + \left(\sum_{k=i}^{m+i-1}W(k,n+j-1) - \sum_{i=1}^{m}T(i,n)\right)^2}$$
(7)

and

$$C(i,j) = \left|\sum_{k=i}^{m+i-1}W(k,j) - \sum_{i=1}^{m}T(i,1)\right| + \left|\sum_{k=i}^{m+i-1}W(k,j+1) - \sum_{i=1}^{m}T(i,2)\right| \ldots + \left|\sum_{k=i}^{m+i-1}W(k,n+j-1) - \sum_{i=1}^{m}T(i,n)\right|$$
(8)

Figs. 4 and 5 compare between the three similarity functions for the images greens and lifting-body, respectively. In this comparing we taken five template with different sizes from each reference and computing the time confused by each similarity function in our proposed method. As can be seen from Fig. 4, SSD function outperforms Euclidean and SAD functions but the required time of the three functions is almost the same at template of sizes 25×25 and 150×150. Also, from Fig. 5, we can see that SSD function outperform Euclidean and SAD functions but the required time for the three functions is almost the same at template of sizes 25×25 and 200×200. Finally, from these two figures, we can see that for the two kind of images the SSD functions is the best because the number of operations of SSD is less than the number of operations in SAD and Euclidean (see Eqs. (3), (7) and (8)).

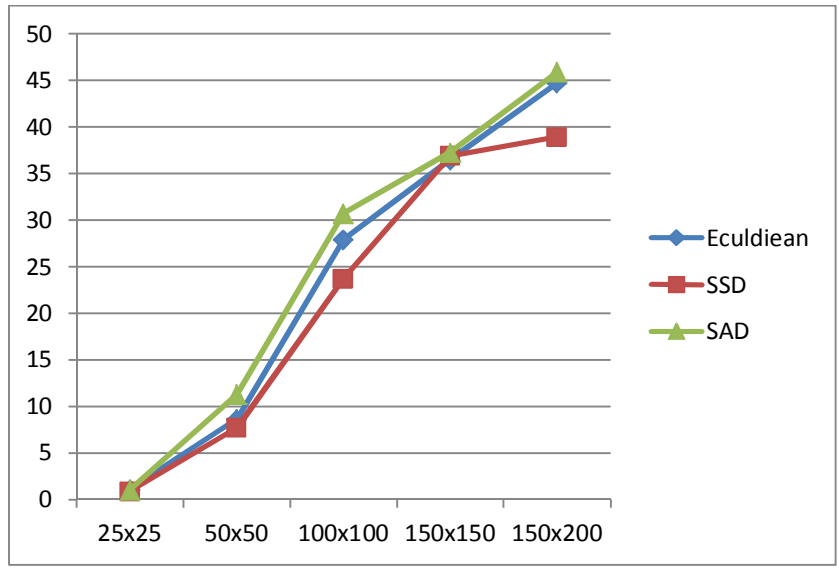

Fig. 4. Comparisons between the three similarity functions on greens image using different size of template

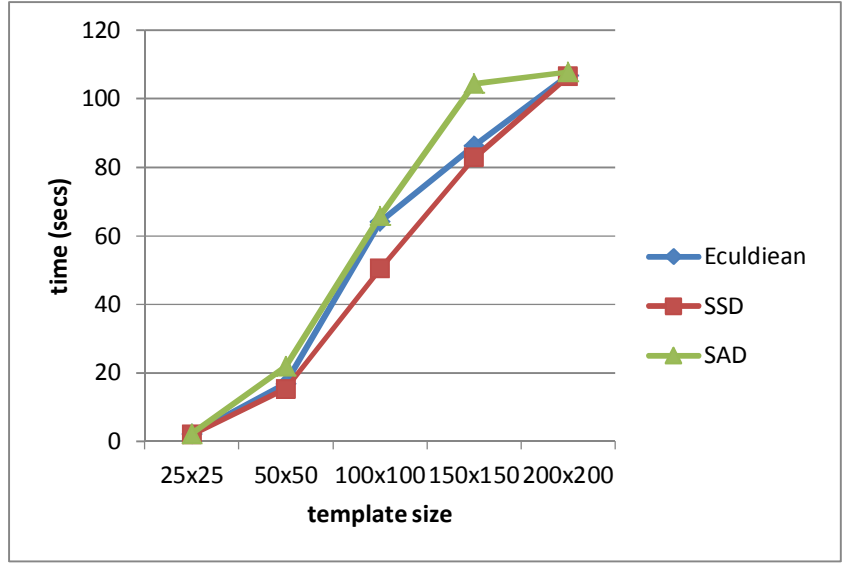

Fig. 5. Comparisons between the three similarity functions on lifting-body image using different size of template

The execution time required for the proposed and compared algorithms is shown in Tables I and II. These Tables shows the performance comparison of the above mentioned algorithms using color image of size 300×500 and gray scale image of size 512×512 respectively. For each reference image the templates are cropped with sizes from 25×25 to 200×200. From these Tables we see that the times in Table I is less than its corresponding times in Table II because the size of reference image in Tables I and II are 150000 and 262141 pixels respectively. In Table I the running time for NCCP and SADP algorithms for 150×150 template is greater than 200×200 template although the converse must be satisfied. This result because the 150×150 template is very close the y-direction in greens image. NCCP and SADP give a better results than NCC and SAD when the template is far from the x-direction and y-direction in the reference image.

Table I Execution time by seconds of applying NCC, NCCP, SAD, SADP, and proposed method with five templates shown in fig. 2(b) and the reference image show in fig. 2(a).

| Template size | Algorithm | | | | |
| --- | --- | --- | --- | --- | --- |
| | NCC | NCCP | SAD | SADP | Proposed |
| 25×25 | 18.95 | 8.97 | 5.12 | 3.03 | 1.06 |
| 50×50 | 56.58 | 42.25 | 16.66 | 14.19 | 7.7 |
| 100×100 | 158.28 | 75.16 | 45.87 | 24.78 | 23.17 |
| 150×150 | 223.52 | 135.42 | 68.56 | 46.05 | 36.88 |
| 150×200 | 255.05 | 121.17 | 78.92 | 41.52 | 38.92 |

Table II Execution time by seconds of applying NCC, NCCP, SAD, SADP, and proposed method with five templates shown in fig. 3(b) and the reference image show in fig. 3(a).

| Template size | Algorithm | | | | |
| --- | --- | --- | --- | --- | --- |
| | NCC | NCCP | SAD | SADP | Proposed |
| 25×25 | 35.52 | 21.26 | 10.31 | 6.81 | 2.26 |
| 50×50 | 120.26 | 49.86 | 34.4 | 17.38 | 15.23 |
| 100×100 | 339.05 | 213.23 | 104.19 | 75.12 | 50.39 |
| 150×150 | 573.36 | 290.14 | 198.52 | 103.45 | 94.86 |
| 200×200 | 779.91 | 349.09 | 241.12 | 129.27 | 106.6 |

From table I, it is clarified that the processing time for the proposed is less than that for NCC, NCCP, SAD, and SADP. For example, using the 50×50 template, the processing time for the proposed was 7.7 s, however, NCC, NCCP, SAD, and SADP matching required 56.58, 42.25, 16.66, and 14.19 s respectively. This means that the proposed give an improvement at least about 50%. Also from table II the required time of the proposed is better than another all algorithms for all templates. Since the complexity of the proposed is more efficient than the other methods. Figs. 6 and 7 shows the performance of the proposed algorithm compared with other four algorithms using greens image (color case) and lifting-body (gray scale case) respectively. It is clear that the proposed algorithm outperform the other in the two cases for all template sizes. In the above experiments, the correct match position is assumed to be the position where the minimum similarity distance value is obtained when the entire template is used in the search process.

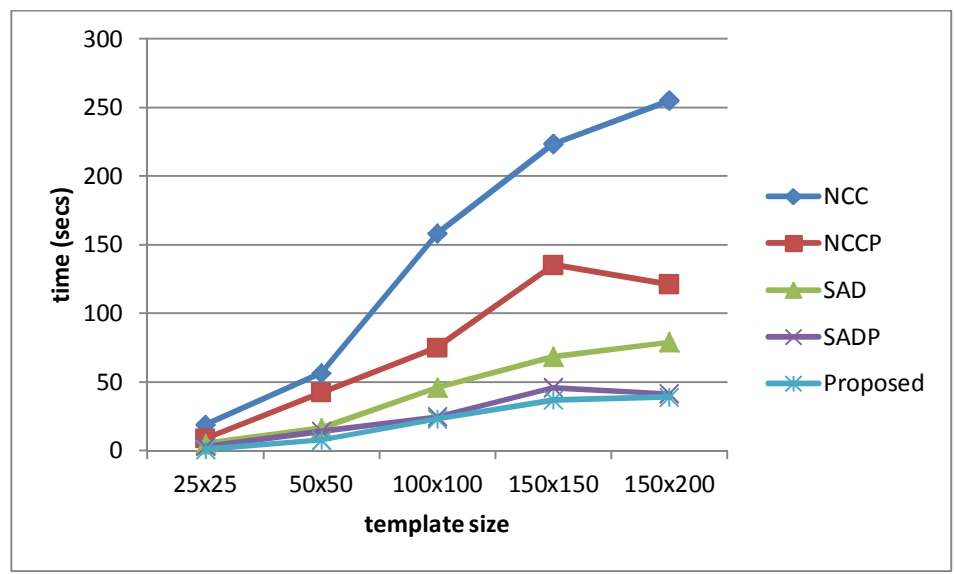

Fig. 6. Performance of the proposed algorithm using greens image

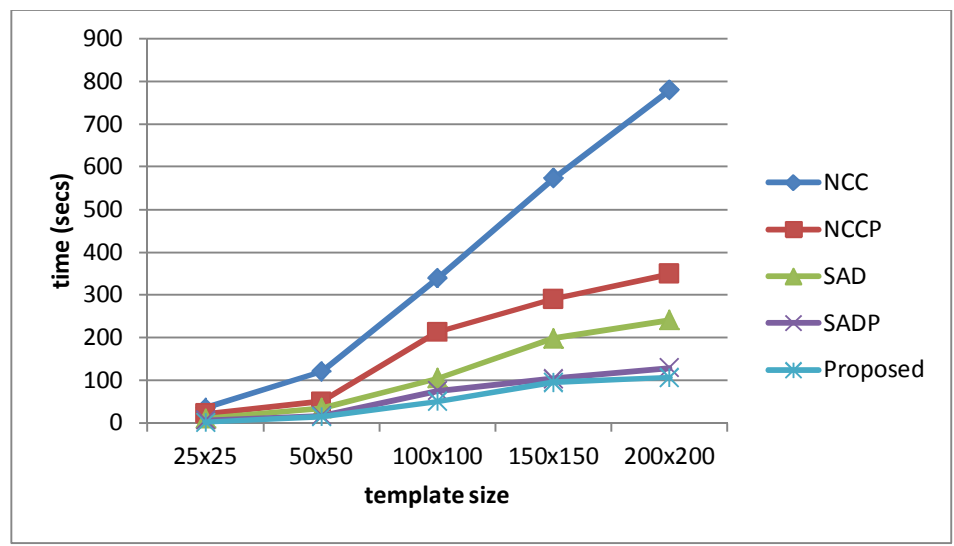

Fig. 7. Performance of the proposed algorithm using lifting-body image

## 4 Conclusion

In this paper, we have proposed a new template matching which can be speed up the computation of block matching while still guaranteeing the correct match for template in the reference. To achieve efficient computation, we converted the template and each corresponding block in the reference from 2-D into 1-D. We have applied the proposed idea to the template matching using three different similarity measure, and have a reduction in computation time. Two different types of image (color and gray scale) are used for comparison between proposed algorithm and other algorithms. The templates are cropped from the reference image. The experimental results show the proposed algorithm is efficient for pattern matching under uniform illumination.